\documentclass[runningheads]{llncs}
\usepackage[T1]{fontenc}
\usepackage{graphicx}
\usepackage{booktabs}
\usepackage{amssymb}
\usepackage{amsmath}
\usepackage[misc]{ifsym}
\usepackage{multirow}
\usepackage{array}
\usepackage{xcolor}
\usepackage{amsfonts}
\usepackage{bbm}
\usepackage{xurl}
\usepackage{bm}
\usepackage{float}

\newcommand{\corr}{(\Letter)}
\usepackage{mwe}

\begin{document}

\title{The Role of Hyperparameters in Predictive Multiplicity}

\titlerunning{The Role of Hyperparameters in Predictive Multiplicity}

\author{Mustafa Cavus\inst{1}\corr\and
Katarzyna Woźnica\inst{2}\and
Przemysław Biecek\inst{2,3}}

\authorrunning{Cavus et al.}

\institute{Eskisehir Technical University, Department of Statistics, Turkiye \email{mustafacavus@eskisehir.edu.tr}
\and Faculty of Mathematics and Information Science, Warsaw University of Technology, Poland
\and Faculty of Mathematics, Informatics and Mechanics, University of Warsaw, Poland}

\maketitle             

\begin{abstract}
This paper investigates the critical role of hyperparameters in predictive multiplicity, where different machine learning models trained on the same dataset yield divergent predictions for identical inputs. These inconsistencies can seriously impact high-stakes decisions such as credit assessments, hiring, and medical diagnoses. Focusing on six widely used models for tabular data—Elastic Net, Decision Tree, k-Nearest Neighbor, Support Vector Machine, Random Forests, and Extreme Gradient Boosting—we explore how hyperparameter tuning influences predictive multiplicity, as expressed by the distribution of prediction discrepancies across benchmark datasets. Key hyperparameters such as \texttt{lambda} in Elastic Net, \texttt{gamma} in Support Vector Machines, and \texttt{alpha} in Extreme Gradient Boosting play a crucial role in shaping predictive multiplicity, often compromising the stability of predictions within specific algorithms. Our experiments on 21 benchmark datasets reveal that tuning these hyperparameters leads to notable performance improvements but also increases prediction discrepancies, with Extreme Gradient Boosting exhibiting the highest discrepancy and substantial prediction instability. This highlights the trade-off between performance optimization and prediction consistency, raising concerns about the risk of arbitrary predictions. These findings provide insight into how hyperparameter optimization leads to predictive multiplicity. While predictive multiplicity allows prioritizing domain-specific objectives such as fairness and reduces reliance on a single model, it also complicates decision-making, potentially leading to arbitrary or unjustified outcomes.

\end{abstract}
\section{Introduction}
\label{sec:intro}

Training machine learning models typically involves three main stages: dataset preparation, model selection and training, and model evaluation. Each stage impacts the model's performance, but a critical factor often overlooked in the process is how the model behaves under different configurations. These behaviors can lead to different models trained on the same dataset producing distinct outputs for the same inputs. This phenomenon is referred to as predictive multiplicity \cite{marx_et_al_2020}, also known as prediction discrepancy \cite{renard_et_al_2024} or prediction disagreement \cite{roth_et_al_2023,du_et_al_2024}, which describes the phenomenon where different or differently configured models trained on the same dataset yield distinct outputs for identical inputs. This variability can introduce significant inconsistencies, especially in high-risk decisions such as credit assessment, hiring, and medical diagnoses, where individuals may face drastically different results depending on which model is used \cite{long_et_al_2023}. Therefore, predictive multiplicity raises serious concerns regarding fairness, accountability, trustworthiness, and the reliability of machine learning-based decisions.

One of the key factors contributing to predictive multiplicity is the inherent uncertainty in the dataset and the randomness introduced during the training process \cite{renard_et_al_2024}. Issues such as missing or imbalanced data, sampling errors, and the model's ability to learn patterns can contribute to this phenomenon. Additionally, different choices in model design, such as variations in training algorithms, loss functions, hyperparameter choices, and data preprocessing methods, can directly influence predictive multiplicity \cite{cavus2024investigating,gomez_et_al_2024}. Arbitrary decisions made during training may disproportionately affect certain input samples, leading to unfair treatment \cite{long_et_al_2023,cooper_et_al_2024}. Studies have shown that hyperparameters like learning rates and regularization parameters can significantly impact model predictions, further contributing to the variability in outputs, which is often overlooked in conventional evaluation processes \cite{cavus2024experimental,renard_et_al_2024,ganesh_et_al_2024}.

Addressing predictive multiplicity is essential due to its potential consequences at the individual and societal levels. For example, \cite{long_et_al_2023} highlights that even models designed to ensure group fairness may still exhibit arbitrariness in their predictions, which can silently harm individuals. While multiplicity can increase model robustness by offering flexibility, this inherent arbitrariness may undermine fairness and transparency \cite{watson-daniels_et_al_2023,sokol_et_al_2024}.

Despite growing recognition of predictive multiplicity, the role of hyperparameter tuning in shaping this issue remains underexplored. While prior work examines hyperparameters' effects on model performance \cite{van_et_al_2018,probst_et_al_2019}, post-hoc explanations \cite{muller_et_al_2023,wickstrom_et_al_2024}, and their interplay with performance \cite{grushetskaya_et_al_2024}, their specific impact on predictive multiplicity has been largely overlooked. Hyperparameter optimization traditionally selects the best model based on a single performance metric~\cite{bischl_2021}, whereas multiobjective optimization broadens evaluation to factors like training or inference time~\cite{karl_2023}. However, these methods still prioritize overall model performance, disregarding the ambiguity in individual predictions. Similarly, research on the optimization process primarily assesses model quality~\cite{grushetskaya_et_al_2024,sass_2022,golovin_2017,wang_2019} using metrics like AUC or accuracy, reinforcing a narrow focus on single-model selection. Even AutoML frameworks, which aim to generalize across diverse datasets using model portfolios~\cite{wistuba_2015a,feurer_2022} or multiple defaults~\cite{pfisterer_2021a}, fail to address predictive multiplicity within a single dataset.



Understanding how hyperparameters affect model decision-making and contribute to predictive multiplicity could provide valuable insights into improving fairness and transparency in machine learning systems. For example, \cite{hsu2024dropout} emphasizes that ignoring multiplicity in model selection may lead to systematic exclusion or discriminatory practices.

In this paper, we aim to examine the effect of hyperparameter tuning on predictive multiplicity, providing both theoretical and experimental approaches to measure how variations in hyperparameters influence model predictions. By bridging this gap in the existing literature, we aim to contribute to the understanding of predictive multiplicity and support ongoing discussions about the reliability of machine learning models. Our findings will help model developers control the effects of hyperparameters on predictive multiplicity when tuning them. Moreover, we believe that being aware of this effect can lead to encouraging the integrant modeling approaches, such as the Rashomon perspective \cite{breiman_2001,biecek_et_al_2024}, to model the data in various ways \cite{rudin_et_al_2024}.

The paper is structured as follows: Section~\ref{sec:back} defines theoretical measures for model, hyperparameter, and joint multiplicity. Section~\ref{sec:exp} covers the experimental design and datasets. Results are in Section~\ref{sec:res}, followed by discussion in Section~\ref{sec:dis} and conclusion in Section~\ref{sec:conc}.
\section{Background}
\label{sec:back}

Let $D = \{(\mathbf{x}_i, y_i)\}_{i=1}^n$ represent a dataset consisting of $n$ observations across $p$ variables, where $\mathbf{x}_i = [x_{i1}, x_{i2}, \ldots, x_{ip}] \in \mathcal{X}$ is the predictor vector, and $y_i \in \mathcal{Y}$ is the corresponding response. Assume that the pairs $(X, Y)$ follow an unknown distribution $\mathcal{P}(\mathcal{X}, \mathcal{Y})$. The objective is to identify the best model $f^*: \mathcal{X} \to \mathcal{Y}$ belonging to the model family $ \mathcal{F}= \{ f_{\bm{\theta}} \}$, parameterized by hyperparameter configuration $\bm{\theta} = (\theta^{[1]}, \theta^{[2]}, ..., \theta^{[k]})$ from the $k$-dimensional hyperparameter search space $\bm{\Theta} = \Theta^{[1]} \times \Theta^{[2]} \times \Theta^{[k]}$.  Formally, model $f^*$ minimizes the expected loss:
\begin{equation}
    f^* = \underset{f_{\bm{\theta}} \in \mathcal{F}}{\arg\min}\, \mathbb{E}_{(X,Y) \sim P} [L(Y, f_{\bm{\theta}}(X))],
\end{equation}

\noindent where $L: \mathcal{Y} \times \mathcal{Y} \to \mathbb{R}$ is a loss function that quantifies the model's error.

In the following sections, we define the theoretical measures for the discrepancy of a model, the discrepancy of a hyperparameter, and the joint discrepancy of hyperparameter combinations.

\subsection{Measuring Discrepancy of a Model}
\label{sec:mmm}

The predictive multiplicity (hereafter briefly referred to as \textit{discrepancy}) of a model trained on a dataset across $k$ hyperparameter configurations can be computed based on the \textit{discrepancy} metric \cite{marx_et_al_2020} --- the proportion of conflicting predictions --- between the predictions of a model trained with default hyperparameters and the predictions of the same model trained with tuned hyperparameters. Let $f_{\bm{\theta}_j}$ denote the models trained with hyperparameter configuration $\bm{\theta}_j \in \bm{\bar{\theta}} = \left\{\bm{\theta}_1, \bm{\theta}_2, \dots, \bm{\theta}_{k} \right\}$. Let $f_{\bm{\theta}_d}$ denote the model trained with the default hyperparameter configuration $\bm{\theta}_d$. The discrepancy of a model trained on a dataset $D_p$ is given by:
\begin{equation}
    \label{eq:mult_of_a_model}
    \delta^{D_p}_{\bm{\bar{\theta}}}(f) = \max_{\bm{\theta}_j \in \bm{\bar{\theta}} /\bm{\theta}_d} \Big( \frac{1}{n} \sum_{i=1}^{n} \mathbbm{1} [f_{\bm{\theta}_j}(x_i) \neq f_{\bm{\theta}_d}(x_i)]\Big),
\end{equation}

\noindent where $n$ is the number of observations in dataset $D_p$ and $x_i$ are its instances. For readability, we omit the $D_p$ symbol in the following sections, but by default this $\delta$  is used for a specific dataset.  For a given set of hyperparameters $\bm{\bar{\theta}}$,  $\delta_{\bm{\bar{\theta}}}(f)$ returns the maximum discrepancy of the hyperparameter configurations. For $m$ datasets, an empirical distribution of the discrepancy of a model can be obtained and summarized using a summary statistic as follows:
\begin{equation}
    g(\delta^{D_l}_{\bm{\bar{\theta}}}(f)), \quad l = 1, 2, \dots, m,
\end{equation}

\noindent where $g(.)$ can be a summary function such as the mean, median, or other statistic of choice.

\subsubsection{Selection of baseline (default) model.}
Originally, the definition of discrepancy uses the baseline model that minimizes the specified loss function as a reference~\cite{marx_et_al_2020}. This avoids the problem of referring to the prediction of a model that is sub-optimal. However, for new data, finding such a model is time-consuming and requires a lot of model generalization validation. For this reason, we decided to treat a trained model with default hyperparameters as the baseline.

\subsection{Measuring Discrepancy of a Hyperparameter}

The discrepancy of a hyperparameter $h$ on a dataset can be computed as the discrepancy between the predictions of a model trained with default hyperparameters and the predictions of the same model trained with the hyperparameters when the remaining hyperparameters are set to default. Let $f_{\theta_j^{[h]}}$ denote the models trained with the hyperparameters when all other hyperparameters are set to default $\theta_j^{[h]}$, and let $f_{\theta_d}$ denote the model trained with the default hyperparameter configuration $\theta_d$. The discrepancy of a hyperparameter on a dataset is given by:

\begin{equation}
\label{eq:mult_of_a_hyp}
    \delta_{\bm{\bar{\theta}}^{[h]}}(f) = \max_{\bm{\theta}_j \in \bm{\bar{\theta}}, \theta_j^{[k]} =\theta_d^{[k]} ; \forall k \neq h  } \Big( \frac{1}{n} \sum_{i=1}^{n} \mathbbm{1} [f_{\bm{\theta}_j}(x_i) \neq f_{\bm{\theta}_d}(x_i)]\Big).
\end{equation}

\noindent For $m$ datasets, an empirical distribution of the discrepancy of a hyperparameter can be obtained and summarized using a summary statistic as in Section~\ref{sec:mmm}.

\subsection{Joint Discrepancy of Hyperparameter Combinations}

Consider two hyperparameters $h_1$ and $h_2$, to compute the joint discrepancy of these hyperparameters on a dataset is the discrepancy between the predictions of a model trained with default hyperparameters and the predictions of the same model trained with these hyperparameters. In contrast, the remaining hyperparameters are set to default. Let $f_{\theta_j^{[h_1, h_2]}}$ denote the models trained with the considered hyperparameters when all hyperparameters except $h_1, h_2$ are set to corresponding values from default configuration $\theta_d$.
The joint discrepancy of the considered hyperparameters on a dataset is given by:
\begin{equation}
\label{eq:mult_of_joint}
    \delta_{\bm{\bar{\theta}}^{[h_1, h_2]}}(f) = \max_{\bm{\theta}_j \in \bm{\bar{\theta}}, \theta_j^{[k]} =\theta_d^{[k]} ; \forall k  \notin \{h_1, h_2\}  } \Big( \frac{1}{n} \sum_{i=1}^{n} \mathbbm{1} [f_{\bm{\theta}_j}(x_i) \neq f_{\bm{\theta}_d}(x_i)]\Big).
\end{equation}

\noindent For $m$ datasets, an empirical distribution of the joint discrepancy of these hyperparameters can be obtained and summarized using a summary statistic as in Section~\ref{sec:mmm}.


\subsection{Measuring Tunability of a Hyperparameter}
While the discrepancy measures how much the model predictions vary depending on the value of a specific hyperparameter, tunability measures how a specific hyperparameter affects model performance.

In this paper, we adopt a modification of the definition of tunability presented in \cite{probst_et_al_2019}, defining it as the F1-score gain from tuning compared to defaults, i.e.

\begin{equation}
\label{eq:tunability}
    d(f_{\theta^{(h)}}) = \max_{j \in \theta^{(h)}/\theta_d} \Big( 
    F_1(f_j) - F_1(f_{\theta_d})\Big).
\end{equation}

\section{Experiments}
\label{sec:exp}

We used models Elastic Net \cite{elasticnet}, Decision Tree \cite{decisiontree}, k-Nearest Neighbors \cite{knn}, Support Vector Machine \cite{svm}, Random Forest \cite{rf}, and Extreme Gradient Boosting \cite{xgb}. The following sections detail datasets and modeling. 

\subsection{Datasets}

We used a benchmark dataset \cite{stando_et_al_2024} to examine hyperparameters' impact on performance and predictive multiplicity. It includes multiple datasets with at least 1,000 observations and 6 to 121 variables for binary classification. 

\subsection{Modeling}

We train the models considered with varying hyperparameters, as shown in Table~\ref{tab:design}, using the \texttt{mlr3} framework \cite{mlr3}. For performance estimation, we use F1-scores due to the imbalanced distribution of the target variable's classes. The number of trials for each hyperparameter configuration depends on the number of hyperparameters in the respective model. The total number of hyperparameter configurations for each model across 21 datasets is approximately 500,000, except for k-Nearest Neighbors.

\section{Results}
\label{sec:res}

The results of the experiments are presented in the following sections.

\subsection{Discrepancy of the models}

\begin{figure}[H]
    \centering
    \includegraphics[width = 0.9\linewidth]{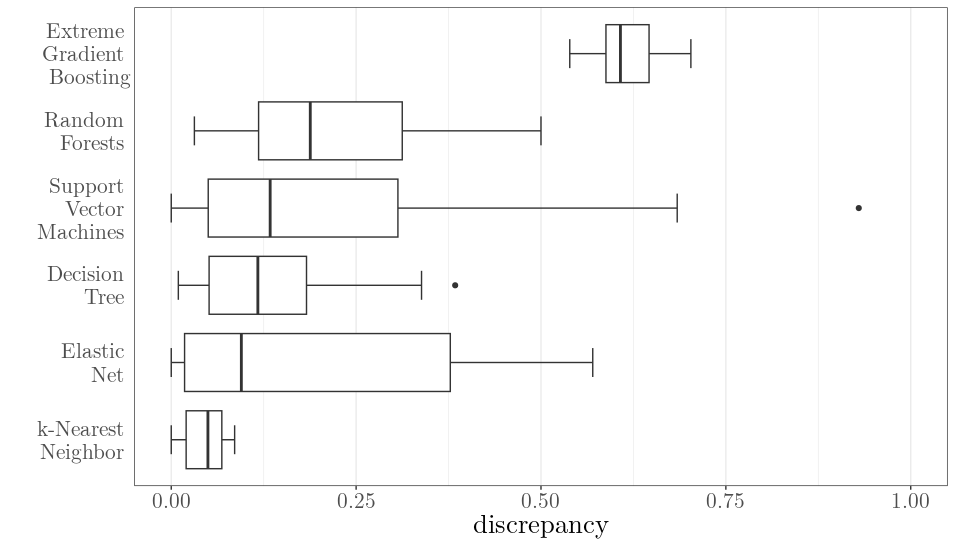}
    \caption{The distribution of the predictive multiplicity of the models in terms of \textit{discrepancy} for the defaults. The discrepancy per model is calculated as the maximum difference between a model trained on the default and tuned hyperparameters on each dataset as in Equation~\ref{eq:mult_of_a_model}.}
    \label{fig1}
\end{figure}

The distribution of the discrepancy values between the models trained on default and tuned hyperparameters is given in Figure~\ref{fig1}. k-Nearest Neighbor (kNN) exhibits the lowest discrepancy with a narrow distribution, suggesting consistent predictions. Decision Tree (DT) also shows relatively low discrepancy, though with a few outliers. Elastic Net (EN) and Support Vector Machines (SVM) display high variability, with wide interquartile ranges and long whiskers, indicating that their predictions can be unstable for some datasets. Random Forests (RF) have a moderate discrepancy level but include some extreme values. Extreme Gradient Boosting (XGB) presents the highest discrepancy, yet with a compact distribution. These findings suggest that kNN or DT may be preferable for less discrepancy, whereas EN and SVM exhibit greater discrepancy.

\subsection{Discrepancy of the hyperparameter of the models}
\label{sec:mult_of_hyp_of_models}

\begin{figure}[H]
    \centering
    \includegraphics[width = 0.9\linewidth]{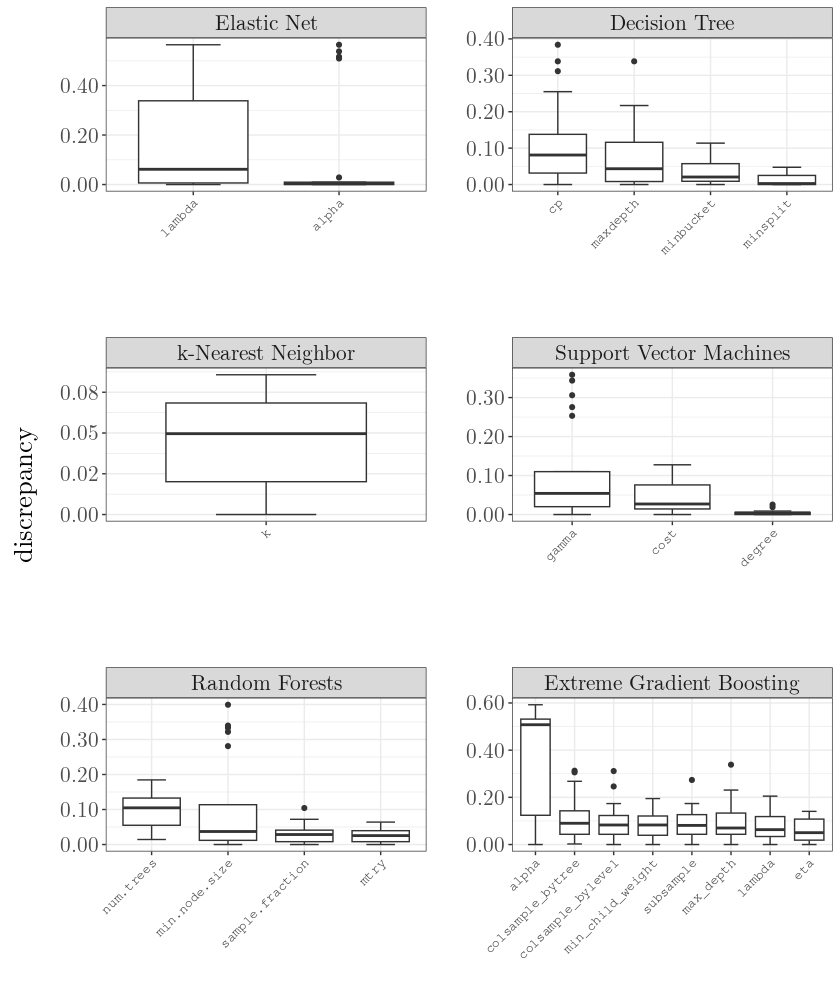}
    \caption{The distribution of the discrepancy of the hyperparameters of the models. The discrepancy per hyperparameters is calculated as the maximum difference between a model trained on the other hyperparameter values, which are fixed to their default values, and the all considered values of the interested hyperparameter on each dataset as in Equation~\ref{eq:mult_of_a_hyp}.}
    \label{fig2}
\end{figure}

The distribution of hyperparameter discrepancy values is shown in Figure~\ref{fig2}. The \texttt{lambda} parameter of EN has high discrepancy, indicating that stronger regularization leads to greater prediction variability, whereas \texttt{alpha} shows consistently low discrepancy, suggesting it primarily affects coefficient values for a given \texttt{lambda}. The variability in kNN stems entirely from the number of \texttt{k} neighbors, though this effect is minor, as kNN is a stable algorithm in terms of discrepancy. In Random Forests, \texttt{num.trees} and \texttt{min.node.size} exhibit relatively high discrepancy, implying they influence prediction stability, while \texttt{sample.fraction} and \texttt{mtry} show lower discrepancy, suggesting more stable results.

The \texttt{cp} and \texttt{maxdepth} hyperparameters of DT introduce notable variability, suggesting that pruning and tree depth may affect model predictions. In contrast, \texttt{minbucket} and \texttt{minsplit} hyperparameters appear to have a lesser impact. SVM exhibits substantial discrepancy in the \texttt{gamma} hyperparameter, meaning that variations in the \texttt{degree} influence can lead to considerable prediction discrepancy, whereas \texttt{cost} has moderate discrepancy and \texttt{degree} remains relatively stable. XGB reveals the highest discrepancy for \texttt{alpha}, showing that regularization strength plays a crucial role in prediction differences, while other hyperparameters \texttt{lambda}, \texttt{max\_depth}, \texttt{eta}, and \texttt{subsample} have similar and relatively lower discrepancy values, suggesting they contribute less to prediction instability.

Overall, the results indicate that \texttt{lambda} in EN, \texttt{gamma} in SVM, \texttt{cp} and \texttt{maxdepth} in DT, and \texttt{alpha} in XGB lead to substantial prediction variability, making them critical tuning hyperparameters in terms of stability of predictions. On the other hand, \texttt{alpha} in EN, \texttt{k} in kNN, \texttt{degree} in SVM, and \texttt{minsplit} in DT exhibit more stability. These findings suggest that careful tuning of high-discrepancy parameters is essential when aiming for consistent model predictions.

\subsection{Discrepancy and tunability of the models}

The results in Table~\ref{tab:mult_tun} highlight the trade-off between discrepancy and tunability \cite{probst_et_al_2019} which is defined as the differences between the performance of the model tuned by the default hyperparameters and when the hyperparameters are set to the optimal values across models. We measure the tunability of the models in terms of \textit{performance gain by F1-score} as in Equation~\ref{eq:tunability}. XGB shows the highest discrepancy, reflecting significant prediction variability due to the regularization parameter \texttt{alpha} (Section~\ref{sec:mult_of_hyp_of_models}). This suggests a risk of arbitrary predictions. However, XGB also exhibits high tunability, indicating notable performance gains with moderate dataset variability.

\setlength{\tabcolsep}{8pt}
\begin{table}[H]
    \centering
    \caption{Summary statistics of model discrepancy and tunability are provided. Discrepancy is the max difference between a model with default and tuned hyperparameters (Equation~\ref{eq:mult_of_a_model}), while tunability is the F1-score gain from tuning compared to defaults, as in Equation~\ref{eq:tunability}. Results are shown as mean $\pm$ standard deviation of discrepancy and F1-score gain across datasets.}
    \label{tab:mult_tun}
    \begin{tabular}{lcc}\toprule
        \textbf{Model}              & \textbf{discrepancy}& \textbf{tunability}           \\\midrule
        Elastic Net                 & 0.2020 $\pm$ 0.2170 & 0.0725 $\pm$ 0.1330     \\
        Decision Tree               & 0.1390 $\pm$ 0.1110 & 0.0109 $\pm$ 0.0106     \\
        k-Nearest Neighbor          & 0.0444 $\pm$ 0.0277 & 0.0098 $\pm$ 0.0113     \\
        Support Vector Machine      & 0.2290 $\pm$ 0.2590 & 0.0102 $\pm$ 0.0107     \\
        Random Forests              & 0.2100 $\pm$ 0.1260 & 0.0118 $\pm$ 0.0095     \\
        Extreme Gradient Boosting   & 0.6150 $\pm$ 0.0452 & 0.0268 $\pm$ 0.0256     \\\bottomrule
    \end{tabular}
\end{table}

Conversely, kNN has the lowest discrepancy, ensuring stable predictions, but also the lowest tunability, limiting performance improvements. SVM, RF, and EN have similar discrepancies with wide standard deviations, likely due to hyperparameters like \texttt{gamma} and \texttt{lambda}. Their tunability is moderate, but EN shows high variability, implying inconsistent gains. DT offers stable predictions with minimal tuning benefits.

These results emphasize the importance of model-specific tuning to balance performance gains and prediction consistency, especially for high-discrepancy models like XGB and SVM, where careful hyperparameter optimization is crucial.

Note that both hyperparameters (directly) and predictive multiplicity (indirectly) can be associated with model performance. To better understand this relationship, we performed experiments to determine how predictive multiplicity varies with the difference in performance between the default (best) model and models generated with other hyperparameters. In the analyses below, we present results for the performance represented by the F1 measure, but similar results are also obtained for other performance measures of the model. 

\begin{figure}[H]
    \centering
    \includegraphics[width = 0.9\linewidth]{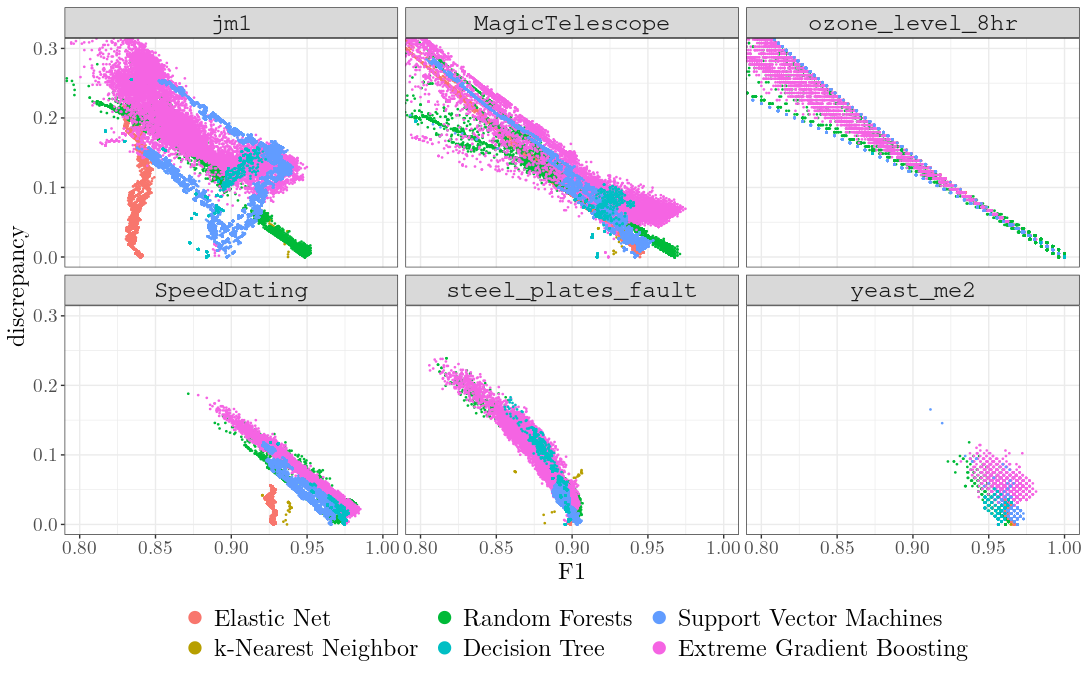}
    \caption{Relationship between predictive multiplicity (measured as discrepancy) and model performance (measured as F1). Each panel shows results for a different dataset. The colors denote the models under consideration. Each dot corresponds to a different set of hyperparameters. Datasets were selected for which variability in discrepancy is observed for models of the same performance.}
    \label{fig4}
\end{figure}

Results from this analysis are presented in Figure~\ref{fig4}. As we may see, models with the same performance (corresponding to the same performance gap) may have significantly different behavior as measured using prediction discrepancy. Both model families differ from each other, as is visible for \texttt{SpeedDating} and \texttt{jm1} data, and models within the same family differ from each other, as is visible for \texttt{yeast\_me} and \texttt{MagicTelescope} data. Flexible model families such as gradient boosting, as expected, also exhibit a greater spectrum of differences among themselves and also differ from models trained with bagging-based techniques such as RF.

This confirms that analyzing the models in terms of performance alone loses important information about the behavior of the models, since even models with the same performance can have this behavior significantly different.

\subsection{Discrepancy of hyperparameter combinations and joint discrepancy}

\begin{figure}[H]
    \centering
    \includegraphics[width = 0.90\linewidth]{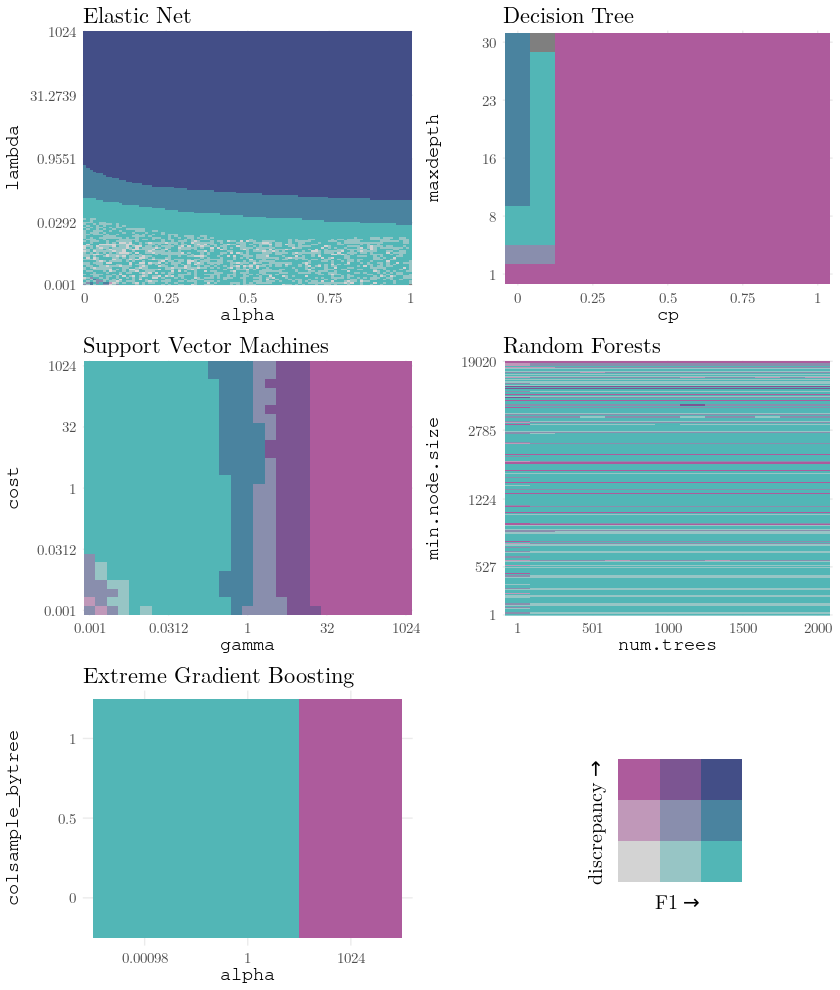}
    \caption{The joint distribution of hyperparameter combinations on discrepancy and model tunability is analyzed. The discrepancy for each hyperparameter joint is calculated as the mean difference between a model trained with default settings for other hyperparameters and one trained with all considered values of the hyperparameter joint, as described in Equation~\ref{eq:mult_of_joint}. We applied a bivariate classification, scaling the mean F1-score and mean discrepancy into a three-by-three grid using the 'equal style,' which divides both variables into three equal-range categories for a balanced distribution. This allows clear visualization of the trade-off between performance and prediction discrepancy in the bivariate heatmaps.}
    \label{fig3}
\end{figure}

The joint distribution of hyperparameter combinations on discrepancy and the tunability of the models is visualized using bivariate heatmaps, as shown in Figure~\ref{fig3}. The discrepancy for each hyperparameter pair is determined by calculating the average difference between a model trained with default settings for all other hyperparameters and a model trained using the full range of considered values for the selected hyperparameter pair, as outlined in Equation~\ref{eq:mult_of_joint}. To analyze the relationship between performance and stability, we applied a bivariate classification approach, where both the mean F1 score and mean discrepancy were scaled into a three-by-three grid. This scaling followed an equal-range method, ensuring that both variables were divided into three balanced categories for a well-distributed classification. 

The colors represent the trade-off between performance and discrepancy: \textcolor[HTML]{d3d3d3}{\textbf{light gray}} tones in the bottom-left corner indicate the lowest values of both F1-score and discrepancy, while moving towards the top-right corner, the colors transition from \textcolor[HTML]{52b6b6}{\textbf{light blue}} to \textcolor[HTML]{434e88}{\textbf{dark blue}} and then to \textcolor[HTML]{ad5b9c}{\textbf{purple}}. This progression reflects an increase in the F1-score along the horizontal axis and a rise in discrepancy along the vertical axis. The darkening of the color tones signifies an increase in the corresponding metric values. Therefore, lighter colors correspond to lower values, whereas darker colors represent higher values.

For EN model, \texttt{alpha} and \texttt{lambda} influence both the discrepancy and F1-score. When \texttt{lambda} is low (below $0.0292$), the F1-score remains relatively high, and discrepancy is minimal. However, as \texttt{lambda} increases beyond $1$, discrepancy rises sharply, leading to greater instability and a decline in performance. This effect is even more pronounced when \texttt{alpha} exceeds $0.5$, suggesting that overly strong regularization negatively impacts the prediction consistency. To achieve better performance with consistent predictions, \texttt{lambda} should be around over $0.0292$, while \texttt{alpha} should be carefully tuned to avoid excessive penalties.

In the DT model, \texttt{cp} and \texttt{maxdepth} play a crucial role in the metrics. When \texttt{cp} is below $0.25$, the F1-score remains relatively stable, though some variations in discrepancy are observed. However, as \texttt{cp} increases beyond this point, the F1-score drops significantly while discrepancy rises, indicating that excessive pruning weakens the model and reduces consistency. Additionally, when \texttt{maxdepth} is shallow (below $8$), F1-scores fluctuate more, suggesting that deeper trees can help capture patterns more effectively—though excessive depth may introduce instability.

For SVM, \texttt{gamma} and \texttt{cost} have a complex impact on discrepancy and F1-scores. At low \texttt{gamma} values (below $0.0312$), the model maintains a stable F1-score with minimal discrepancy. However, when \texttt{gamma} exceeds $1$—especially when \texttt{cost} is also high—discrepancy increases, making the model less reliable. This suggests that overly large \texttt{gamma} values lead to instability in predictions. The F1-score remains higher in regions with lower \texttt{gamma} and moderate \texttt{cost}, reinforcing the importance of avoiding extreme parameter values.

In the RF model, \texttt{num.trees} and \texttt{min.node.size} influence the metrics differently. The number of trees does not significantly affect the F1-score, implying that increasing trees beyond a certain point offers diminishing returns. However, when \texttt{min.node.size} surpasses 1224, discrepancy rises, making the model predictions less consistent. This suggests that an excessively large minimum node size prevents the model from capturing finer details, reducing its predictive reliability.

For XGB model, \texttt{alpha} and \texttt{colsample\_bytree} affect both discrepancy and F1-score. When \texttt{alpha} exceeds $1$, discrepancy increases, leading to less stable predictions and a drop in the F1-score. Conversely, at lower \texttt{alpha} values, discrepancy remains lower while the F1-score improves, indicating a more reliable model. Additionally, using lower \texttt{colsample\_bytree} values appears to enhance F1-scores, suggesting that training on a subset of features helps maintain generalization.

\section{Discussion}
\label{sec:dis}

Hyperparameter tuning plays a crucial role in determining the trade-off between performance improvement and predictive stability. The results from \cite{probst_et_al_2019} indicate that certain hyperparameters significantly enhance model performance when optimized, while our findings reveal that these same hyperparameters often introduce notable predictive discrepancy. This interplay between tunability and predictive multiplicity provides important insights for balancing performance gains and prediction consistency.

For EN, lambda emerges as the most tunable hyperparameter in terms of AUC improvement, whereas \texttt{alpha} contributes comparatively less to performance gains \cite{probst_et_al_2019}. Our results confirm this pattern, showing that \texttt{lambda} exhibits high discrepancy, suggesting that stronger regularization leads to significant prediction variations, while \texttt{alpha} remains more stable. This indicates that while tuning \texttt{lambda} is effective for performance optimization, it may come at the cost of increased model instability.

In DT, \texttt{minbucket} and \texttt{minsplit} were identified as the most impactful hyperparameters for tunability, aligning with \cite{mantovani_et_al_2024}. Our analysis reveals that \texttt{maxdepth} and \texttt{cp} introduce the most predictive discrepancy, suggesting that deeper trees and pruning strategies significantly affect model variability. Notably, while \texttt{min\-bucket} and \texttt{minsplit} influence tunability, they have a lesser impact on discrepancy, indicating that tuning these hyperparameters may provide performance gains without substantially increasing prediction instability.

For SVM, \texttt{gamma} was shown to be highly tunable, yielding significant performance improvements when optimized \cite{probst_et_al_2019}. Our results confirm that \texttt{gamma} also introduces substantial predictive multiplicity, suggesting that while tuning this hyperparameter enhances accuracy, it can lead to unstable decision boundaries that vary across models. This aligns with \cite{van_et_al_2018}, who identified \texttt{gamma} as a primary driver of performance, emphasizing its critical role in SVM behavior. To mitigate this instability, their analysis suggests favoring lower \texttt{gamma} values, which consistently yield robust performance.

In RF, \texttt{mtry} was identified as the most tunable hyperparameter in terms of AUC gains \cite{probst_et_al_2019}. However, our analysis suggests that \texttt{min.node.\-size} and \texttt{num.trees} are more responsible for predictive discrepancy, whereas \texttt{mtry} exhibits lower discrepancy. This indicates that tuning \texttt{mtry} is an effective strategy for enhancing performance while maintaining model stability, whereas \texttt{min.\-node.size} should be adjusted cautiously to avoid excessive prediction variability.

Table~\ref{tab:mult_tun} provides a striking confirmation of this variability for XGB, with a discrepancy—the highest across all models—driven by \texttt{alpha}, as our results indicate. This exceptionally high predictive multiplicity suggests that the regularization strength of \texttt{alpha} consistently introduces significant instability across datasets, posing a severe risk of arbitrary predictions in high-stakes scenarios. Despite this, its F1 tunability indicates a relatively higher performance gain compared to other models, highlighting a critical trade-off. This necessitates advanced tuning strategies, such as constraining \texttt{alpha} to lower values (as in Figure~\ref{fig3}), to balance performance improvements with prediction consistency, ensuring fairness and reliability in high-stake applications.

\section{Conclusion}
\label{sec:conc}

Our findings emphasize the dual need to optimize performance gains while minimizing predictive discrepancy in hyperparameter tuning. The results show that XGB has the highest discrepancy, making it vulnerable to predictive multiplicity, particularly due to \texttt{alpha}, despite notable F1 tunability. In contrast, kNN offers stable predictions but limited tuning benefits.

This trade-off aligns with \cite{grushetskaya_et_al_2024}, which introduces an explainable AI tool to identify performance-consistent subspaces where predictions remain stable despite minor hyperparameter variations. Integrating such tools could help practitioners visually explore these subspaces—e.g., favoring lower \texttt{gamma} in SVM and constraining \texttt{alpha} in XGB—enhancing transparency and reliability. In critical applications like flood prediction or medical diagnostics, where inconsistent outcomes may affect fairness, these strategies are essential for balancing predictive power with accountability.

This variability underscores model-specific effects and broader study limitations: (1) the analysis used 21 imbalanced datasets, limiting generalizability to balanced data; (2) only six ML models were considered, excluding deep learning; (3) the study focused on specific hyperparameters, omitting others that may influence predictive discrepancy; (4) alternative tuning strategies, such as Bayesian optimization, were not explored; and (5) while ethical concerns were noted, further empirical studies are needed to assess real-world impacts.

Future research should examine diverse datasets, including real-world and balanced data, to better understand predictive multiplicity. Exploring alternative tuning methods could further clarify trade-offs between performance and stability, supporting the development of fair and robust ML models.

\begin{credits}
\subsubsection{\ackname} This study is funded by the Polish National Science Centre under SONATA BIS grant 2019/34/E/ST6/00052 and Eskisehir Technical University Scientific Research Projects Commission.

\end{credits}


%
%
%
%

\appendix
\renewcommand{\thetable}{A\arabic{table}} 
\setcounter{table}{0}
\section{Hyperparameter configurations}

\begin{table}[h]
    \centering
    \caption{The \textbf{Lower} and \textbf{Upper} values of the hyperparameters, and the \textbf{Default} values of the hyperparameters used in the modeling.}
    \label{tab:design}
    \resizebox{0.90\linewidth}{!}{
    \begin{tabular}{lllrrr}\toprule
        \textbf{Model}              & \textbf{Hyperparameter}       & \textbf{Class}    & \textbf{Lower}    & \textbf{Upper}    & \textbf{Default}  \\\midrule
        Elastic Net                 & \texttt{alpha}                & numeric           & 0                 & 1                 & 1                 \\
                                    & \texttt{lambda}               & numeric           & $2^{-10}$         & $2^{10}$          & 0                 \\\midrule
        Decision Tree               & \texttt{cp}                   & numeric           & 0                 & 1                 & 0.1               \\
                                    & \texttt{maxdepth}             & integer           & 1                 & 30                & 30                \\
                                    & \texttt{minbucket}            & integer           & 1                 & 60                & 7                 \\
                                    & \texttt{minsplit}             & integer           & 1                 & 60                & 20                \\\midrule
        k-Nearest Neighbor          & \texttt{k}                    & integer           & 1                 & 30                & 7                 \\\midrule
        Support Vector Machine      & \texttt{cost}                 & numeric           & $2^{-10}$         & $2^{10}$          & 1                 \\
                                    & \texttt{gamma}                & numeric           & $2^{-10}$         & $2^{10}$          & $1 / p$           \\
                                    & \texttt{degree}               & integer           & 2                 & 5                 & 3                 \\\midrule
        Random Forests              & \texttt{num.trees}            & integer           & 1                 & 2000              & 500               \\
                                    & \texttt{sample.fraction}      & numeric           & 0.1               & 1                 & 1                 \\
                                    & \texttt{mtry}                 & integer           & 0                 & $p$               & $\sqrt{p}$        \\
                                    & \texttt{min.node.size}        & integer           & 1                 & $n$               & 1                 \\\midrule
        Extreme Gradient Boosting   & \texttt{nrounds}              & integer           & 1                 & 5000              & 500               \\
                                    & \texttt{eta}                  & numeric           & 0                 & 1                 & 0.3               \\
                                    & \texttt{subsample}            & numeric           & 0.1               & 1                 & 1                 \\
                                    & \texttt{max\_depth}           & integer           & 1                 & 15                & 6                 \\
                                    & \texttt{min\_child\_weight}   & numeric           & 1                 & 14                & 1                 \\
                                    & \texttt{colsample\_bytree}    & numeric           & 0                 & 1                 & 1                 \\
                                    & \texttt{colsample\_bylevel}   & numeric           & 0                 & 1                 & 1                 \\
                                    & \texttt{lambda}               & numeric           & $2^{-10}$         & $2^{10}$          & 1                 \\
                                    & \texttt{alpha}                & numeric           & $2^{-10}$         & $2^{10}$          & 0                 \\\bottomrule
                                    \multicolumn{6}{r}{$p$ denotes the number of variables and $n$ represents the number of observations}
    \end{tabular}}
\end{table}

\end{document}